\documentclass{article}
\PassOptionsToPackage{numbers, compress}{natbib}
\usepackage[preprint]{neurips_data_2024}

\usepackage[utf8]{inputenc} 
\usepackage[T1]{fontenc}    
\usepackage{hyperref}       
\usepackage{url}            
\usepackage{booktabs}       
\usepackage{amsfonts}       
\usepackage{nicefrac}       
\usepackage{microtype}      
\usepackage{xcolor}         

\usepackage{bm}
\usepackage{color}
\usepackage{CJK}
\usepackage{algorithmic}
\usepackage{algorithm}
\usepackage{bbding} 
\usepackage{multirow}
\usepackage{amsmath}
\usepackage{balance}
\usepackage{amsfonts}
\usepackage{xspace}
\usepackage{enumitem}
\usepackage{tabularx}
\usepackage{graphicx}
\usepackage{float}

\newcommand{\ie}{\emph{i.e.}\xspace} 
\newcommand{\eg}{\emph{e.g.}\xspace} 
\newcommand{\define}[3]{\vspace{1ex}\noindent{ \textbf{\textsc{Definition {#1}}} (#2): \emph{#3}\vspace{1ex}}
}

\newcommand{\ours}{STBench\xspace}

\title{STBench: Assessing the Ability of Large Language Models in Spatio-Temporal Analysis}

\author{%
  Wenbin Li$^{1,2}$, \thanks{Corresponding authors.} Di Yao$^{1}$, Ruibo Zhao$^{1,2}$, Wenjie Chen$^{1,2}$, Zijie Xu$^{1,2}$, Chengxue Luo$^{1,2}$,\\
  \textbf{Chang Gong$^{1,2}$, Quanliang Jing$^{1}$, Haining Tan$^{1}$, $^*$Jingping Bi$^{1}$}\\
  $^1$Institute of Computing Technology, Chinese Academy of Sciences, Beijing, China,\\
  $^2$University of Chinese Academy of Sciences,\\
  \texttt{\{liwenbin20z,yaodi,zhaoruibao23s,chenwenjie23s,xuzijie22s\}@ict.ac.cn} \\
  \texttt{\{gongchang21z,jingquanliang,tanhaining,bjp\}@ict.ac.cn}
}

\begin{document}

\maketitle

\begin{abstract}
  The rapid evolution of large language models (LLMs) holds promise for reforming the methodology of spatio-temporal data mining. However, current works for evaluating the spatio-temporal understanding capability of LLMs are somewhat limited and biased. These works either fail to incorporate the latest language models or only focus on assessing the memorized spatio-temporal knowledge. To address this gap, this paper dissects LLMs' capability of spatio-temporal data into four distinct dimensions: knowledge comprehension, spatio-temporal reasoning, accurate computation, and downstream applications. We curate several natural language question-answer tasks for each category and build the benchmark dataset, namely \ours, containing 13 distinct tasks and over 60,000 QA pairs. Moreover, we have assessed the capabilities of 13 LLMs, such as GPT-4o, Gemma and Mistral. Experimental results reveal that existing LLMs show remarkable performance on knowledge comprehension and spatio-temporal reasoning tasks, with potential for further enhancement on other tasks through in-context learning, chain-of-though prompting, and fine-tuning. The code and datasets of \ours are released on \url{https://github.com/LwbXc/STBench}.
\end{abstract}

\section{Introduction}
The rapid advancement of large language models (LLMs) has opened up new possibilities across various domains~\cite{wang2024survey,thirunavukarasu2023large,abs-2303-18223}. One promising direction is enhancing spatio-temporal data analysis with the ability of LLMs~\citep{abs-2403-00813,LiXX023,abs-2310-06213}. Spatio-temporal data, characterized by both spatial and temporal dimensions, encompasses a variety of datasets crucial for many fields such as geography, meteorology, transportation, and epidemiology. Despite LLMs' remarkable proficiency in language-related tasks, their applicability and effectiveness in handling spatio-temporal data remain relatively unexplored. 

Existing evaluations of spatio-temporal data fall in two categorizes. The first category~\cite{ShiZL22,MirzaeeK22,LiH024a} focus on evaluating the spatial analysis capability of LLMs and design QA pairs of spatial reasoning such as asking "Is the yellow apple to the west of the yellow watermelon?". The QA pairs are constructed in toy environments without temporal information, which is insufficient to assess the ability of LLM on real spatio-temporal tasks. The second category~\cite{abs-2310-02207,abs-2310-14540} aims to evaluate the spatio-temporal analysis capability but only assess the abilities of LLMs' in one specific dimension. For example, the most recent work~\cite{abs-2310-02207} tends to evaluate the memory ability of spatio-temporal knowledge. For a comprehensive evaluation, we argue that the abilities of LLMs in spatio-temporal analysis should contain not only the memory ability but also other dimensions, such as reasoning, inference and knowledge comprehension. 

\begin{figure}
    \centering
    \includegraphics[width=0.9\textwidth]{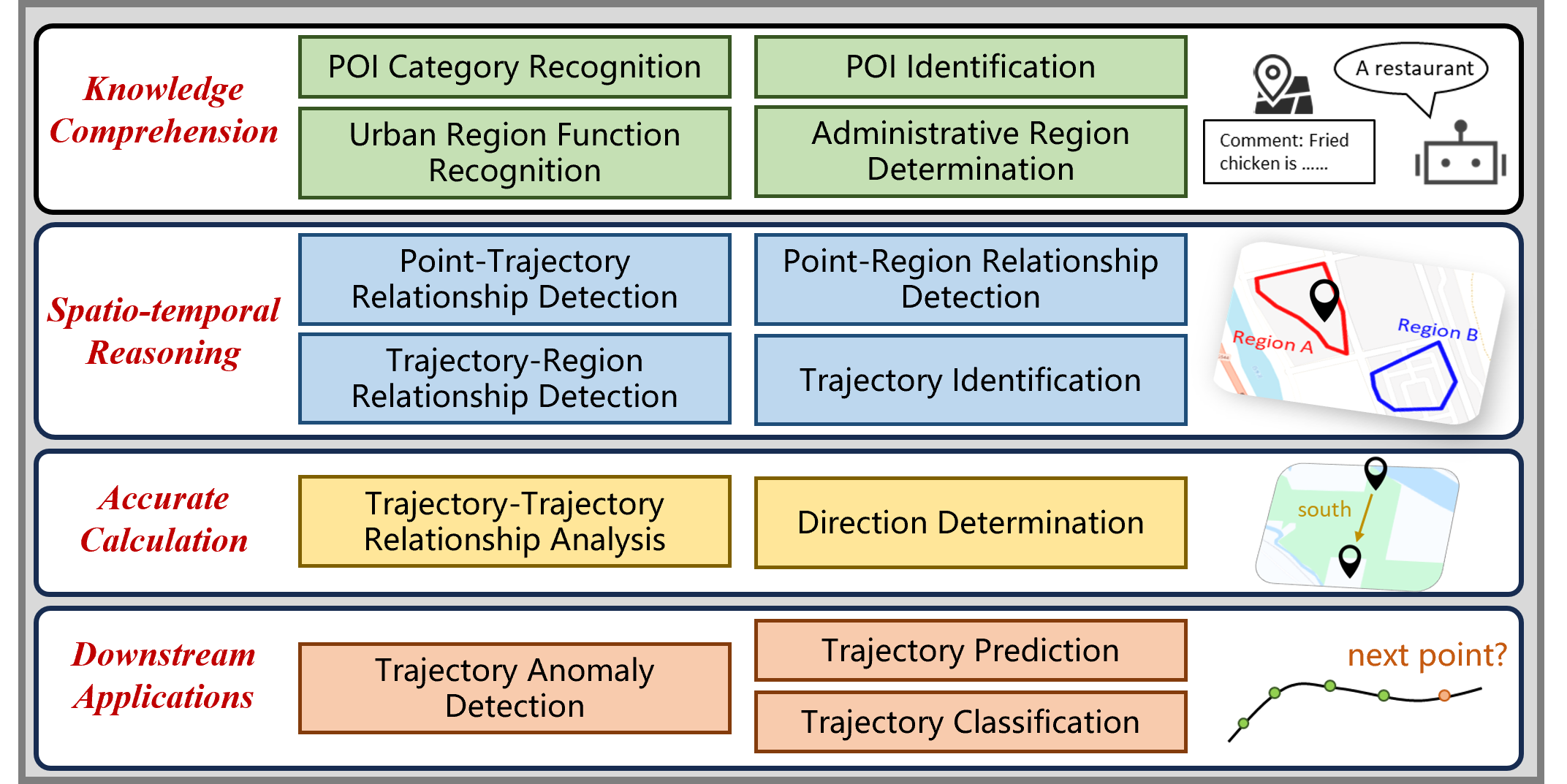}
    \caption{Overview of \ours. It consists of 13 distinct tasks covering four dimensions: knowledge comprehension, spatio-temporal reasoning, accurate calculation and downstream applications.}
    \label{fig:overview}
    \vspace{-3ex}
\end{figure}

To achieve this goal, we propose a framework, namely \ours, for evaluating the spatio-temporal capabilities of LLMs. As shown in Figure~\ref{fig:overview},  \ours dissects the LLMs' capacity into four distinct dimensions: knowledge comprehension, spatio-temporal reasoning, accurate computation, and downstream applications. \textbf{Knowledge Comprehension} examines the model's capacity to understand and interpret the underlying meaning and context of spatio-temporal information. \textbf{Spatio-Temporal Reasoning} evaluates the ability to understand and reason about the spatial and temporal relationships between entities and events. \textbf{Accurate Computation} handles the precise and complex calculations of spatio-temporal data. Moreover, we also employ some \textbf{Downstream Applications} such as trajectory anomaly detection and trajectory prediction to assess the ability of LLMs on practical tasks.

For each evaluated dimension, we design several tasks and construct QA pairs to assess the ability of LLMs qualitatively. We have curated a benchmark dataset, \ours, which contains over 60,000 QA pairs and 13 distinct tasks covering the four dimensions. Furthermore, we evaluated the latest 13 LLMs, including GPT-4o\footnote{https://platform.openai.com/docs/models/gpt-4o}, Gemma~\cite{abs-2403-08295}, Llama2~\cite{abs-2307-09288}, and provide a detailed report that quantitatively assesses the four dimensional abilities of LLMs. Our experimental results reveal that existing LLMs show remarkable performance on knowledge comprehension and spatio-temporal reasoning tasks, with the closed-source LLMs (GPT-4o and ChatGPT\footnote{https://openai.com/blog/chatgpt}) outperforming other models in many instances. For example, ChatGPT achieved an accuracy of 79.26\% on POI Category Recognition and 83.58\% on Administrative Region Determination, surpassing other evaluated open-source models by 34.6\% and 177.3\%, respectively. For accurate computation tasks, performance across all models is generally low. Moreover, we also reveal the potential of in-context learning and chain-of-thought prompting in enhancing performance. For example, in-context learning improved ChatGPT's accuracy on POI Identification from 58.64\% to 76.30\%. Similarly, chain-of-thought prompting increased its accuracy on Urban Region Function Recognition from 39.78\% to 52.20\%. 

The contributions of this paper are summarized as following:
\begin{itemize}
    \item This paper serves as a comprehensive evaluation of many LLMs on spatio-temporal analysis and releases a benchmark dataset \ours. By systematically evaluating their performance across diverse tasks and datasets, we have elucidated the strengths and limitations of LLMs in the context of spatio-temporal analysis.
    \item Our findings highlight the remarkable performance of LLMs in knowledge comprehension and spatio-temporal reasoning tasks, while also identifying areas for improvement in accurate computation and downstream applications. The in-context learning, chain-of-thought prompting, and fine-tuning are verified to be potential techniques in developing more robust and capable models.
    \item For transparency, we have made all the datasets, code, and evaluation methodologies of \ours openly accessible. We believe that sharing our findings and resources will not only facilitate reproducibility but also encourage broader engagement and innovation within the research community. 
\end{itemize}

\section{Related Work}

The rapid development of large-scale language models has attracted widespread interest from various communities~\cite{abs-2402-01801,abs-2310-10196,abs-2402-01749,kasneci2023chatgpt}. Many researchers studied the capabilities of LLMs~\cite{abs-2307-03109,abs-2107-03374,0011LH024} and some of them investigated the potential in spatio-temporal mining.

\textbf{Spatial analysis capabilities.} ~\cite{MirzaeeFNK21} proposed a question-answering (QA) benchmark for spatial reasoning with natural language texts. ~\cite{ShiZL22} presented a QA dataset to evaluate language models' capability of multi-hop spatial reasoning. ~\cite{MirzaeeK22} provided two datasets about spatial question answering and spatial role labeling problems.
~\cite{LiH024a} further improved a previous benchmark to provide a more accurate assessment. However, these works only focus on spatial reasoning in toy environments. They ignore the temporal dimension and are far from the real scenarios of spatio-temporal applications.

\textbf{Spatio-temporal analysis capabilities.} ~\cite{Ji023} evaluated the ability of LLMs to represent geometric shapes and spatial relationships. ~\cite{mooney2023towards} examines the performance of ChatGPT in a geographic information systems exam to evaluate its spatial literacy. ~\cite{abs-2306-00020} investigates the geographic capabilities of GPT-4~\cite{abs-2303-08774} through a series of qualitative and quantitative experiments. ~\cite{abs-2310-02207} analyzes the learned representations of several spatial and temporal datasets by training linear regression probes. ~\cite{abs-2310-14540} evaluates the ability of LLMs to represent and reason about spatial structures, such as squares and hexagons. ~\cite{abs-2401-02404} assesses four closed-source LLMs on a set of tasks, primarily focusing on coding capabilities, such as code interpretation and code generation. These works either only analyze a specific model or only examine the capabilities of a specific aspect, failing to provide a comprehensive evaluation of the latest closed-source and open-source LLMs. The most relevant work is ~\cite{abs-2311-14656} which assesses the geographic and geospatial capabilities of multimodal LLMs. Their tasks are completely designed for multimodal models and are not applicable to single-modal large language models. To comprehensively assess the spatio-temporal ability of LLMs, in this paper, we classify the spatial-temporal abilities into four categories and propose a benchmark consisting of over 60,000 QA pairs based on this. We benchmark 13 latest LLMs to assess their capabilities and to investigate their potential in spatio-temporal mining.
\section{Preliminary}

In spatio-temporal data mining, the concepts of Point of Interest (POI), Trajectory, and Region play a fundamental role in representing and analyzing spatio-temporal data.
Before presenting the construction methodology of our benchmark, we formally define these concepts in this section.

\define{1}{Point of Interest}{A point of interest (POI) is a specific geographic location $\bm{p}=<i_p, lat_p, lon_p, c_p, \mathcal{M}_p>$, where $i_p$ is the ID number, $lat_p$ is the latitude, $lon_p$ is the longitude, $c_p$ denotes the category of this POI and $\mathcal{M}_p=\{m_1, m_2, \cdots\}$ is a set of comments about this POI.}

\define{2}{Trajectory}{Each trajectory $\bm{t}=<t_1, t_2, \dots>$ is a sequence of points, where each point $t_i=<lat_i, lon_i, time_i>$ is a triplet of latitude, longitude and timestamp.}

\define{3}{Region}{A region is a defined area that is distinct from its surroundings. Each region $\bm{r}=<b_r, c_r, \mathcal{P}_r>$ is characterized by its boundary lines $b_r$ and the region function category $c_r$. The set $\mathcal{P}_r=\{p_1, p_2, \cdots\}$ denotes the POIs that fall in this region.} 

\section{Benchmark Construction\label{sec:construction}}

In this section, we propose a benchmark, \ours, to assess the ability of LLMs in spatio-temporal analysis. We will begin by presenting the considerations that guide the design of \ours. Subsequently, we will delve into a detailed exposition of the construction of \ours.

\subsection{Overview}
To construct a benchmark for assessing the ability of 
LLMs in spatio-temporal data, we should first consider the evaluation tasks and the data format.

\textbf{Ability Categories}. Choosing or designing appropriate tasks is crucial for assessing the ability of LLMs in spatio-temporal data mining. Although numerous spatio-temporal tasks, \ie, trajectory anomaly detection and next POI prediction, already exist, they do not provide a comprehensive evaluation of the capabilities in spatio-temporal analysis. We classify the requisite abilities into four categories: \textit{knowledge comprehension}, \textit{spatio-temporal reasoning}, \textit{accurate computation}, and \textit{downstream applications}. For each category, we design several tasks for assessment.

\textbf{Data Format}. Another important question is what data format we should adopt. If we directly ask the model through dialogue and allow open-ended answers, it will bring some problems. Firstly, the response of LLMs is uncontrolled. For instance, models may only apologize for not being able to provide an accurate answer, rather than directly responding to our question. Moreover, open-ended answers make it difficult to identify the final answer of LLMs, \eg, LLMs may reply with a lot of explanation or even some unrelated content. Therefore, we have LLMs complete the input texts, rather than asking LLMs through dialogue. As shown in Table~\ref{tb:data_format}, each data sample in \ours consists of three parts: the question, the options and the guidance. The LLMs should continue the guidance, \ie, they should generate an option number, thus the output is controllable. Note that some LLMs are chat models and do not support text completion, thus we instruct these models to complete the texts through system prompts. The details are in Appendix A in the supplementary material.

\begin{table}
  \caption{A prompt template of the samples in \ours. The {\color{blue}blue} {texts} describe the question. The {\color{brown}brown} texts are the options. The {\color{teal}teal} texts denote the guidance that constrains the output of LLMs.}
  \vspace{-1ex}
  \label{tb:data_format}
  \centering
  \begin{tabularx}{\textwidth}{X}
    \bottomrule
    {\color{blue}Question: Below is the coordinate information and related comments of a point of interest: $\cdots$. Please answer the category of this point of interest.}
    
    {\color{brown}Options: (1) xxxx, (2) xxxx, (3) xxxx, $\cdots$.}

    {\color{teal}Please answer one option.}

    {\color{teal}Answer: The answer is option (} \cr
    \bottomrule
  \end{tabularx}
  \vspace{-3ex}
\end{table}
\subsection{Knowledge comprehension} 

The model's capacity to understand and interpret the underlying meaning and context of spatio-temporal information is important. This involves the ability to comprehend the semantic nuances within the data and the knowledge of relevant spatio-temporal concepts and entities, \eg, understanding and distinguishing different POI categories. We provide valuable insights into LLMs' spatio-temporal knowledge comprehension capabilities through four tasks: POI category recognition, POI identification, urban region function recognition, and administrative region determination.

\textbf{POI Category Recognition (PCR)}.
The semantics of POI are crucial in various applications such as POI recommendation, thus we design this task to evaluate LLM's understanding of POI semantics. Data samples of this task are generated based on the public Yelp dataset\footnote{https://www.yelp.com/dataset.}. Specifically, we randomly sample some POIs from the Yelp dataset for data construction. For each POI $\bm{p}=<i_p, lat_p, lon_p, c_p, \mathcal{M}_p>$, we randomly select two comments $m_{i_1}, m_{i_2}$ from the comment set $\mathcal{M}_p$. Then, LLMs are asked to predict the category $c_p$ of the POI according to its coordinates $<lat_p, lon_p>$ and the selected comments $<m_{i_1}, m_{i_2}>$. The POI category $c_p$ and four other randomly sampled POI categories are provided as options. 

\textbf{POI Identification (PI)}. In this task, the coordinates and comments of two POIs are provided and LLMs are asked to determine if they are the same POI or not. For a POI $\bm{p}=<i_p, lat_p, lon_p, c_p, \mathcal{M}_p>$ in the Yelp dataset, we construct a positive sample (\ie, the answer is "Yes") and a negative sample based on it. For the positive sample, we ask the model if $<lat_p, lon_p, m_{i_1}, m_{i_2}>$ and $<lat_p+\epsilon_{1}, lon_p + \epsilon_2, m_{i_3}, m_{i_4}>$ describe the same POI, where $m_{i_j}, 1\leq j \leq 4$ are comments sampled from the comment set $\mathcal{M}_p$ and $\epsilon_1, \epsilon_2 \sim U(0.0004, 0.0008)$ are minor disturbances to the coordinates. For negative samples, we construct a KD-Tree and sample another POI $\bm{p}'=<i_{p'}, lat_{p'}, lon_{p'}, c_{p'}, \mathcal{M}_{p'}>$ from the nearest five neighbors of $\bm{p}$. Then, the negative sample is constructed based on $<lat_p, lon_p, m_{i_1}, m_{i_2}>$ and $<lat_{p'}, lon_{p'}, m_{i_5}, m_{i_6}>$, where $m_{i_5}, m_{i_6}$ are comments sampled from the comment set $\mathcal{M}_{p'}$.

\textbf{Urban Region Function Recognition (URFR)}. This task requires LLMs to predict the urban region function according to the boundary lines and the POIs located in the region, which evaluates LLMs' understanding of urban regions. To construct data samples, we first match POIs in the Yelp dataset and regions in the New Orleans region dataset\footnote{https://catalog.data.gov/dataset/zoning-district-9939c}, removing POIs that do not fall in any region and regions that contain no more than one POI. After that, for each region $\bm{r}=<b_r, c_r, \mathcal{P}_r>$, we randomly select two POIs $\{\bm{p}_{k} = <i_{p_k}, lat_{p_k}, lon_{p_k}, c_{p_k}, \mathcal{M}_{p_k}>| k=i_1, i_2\}$ from its POI set $\mathcal{P}_r$. For each $p_k$, two comments $m^{p_k}_1, m^{p_k}_2$ are sampled from the comment set $\bm{M}_{p_k}$, where $k\in {i_1,i_2}$. Then, we ask LLMs to predict the region function $c_r$ according to its boundary lines $b_r$, the coordinates and comments of the selected POIs, \ie, $\{<lat_{p_k}, lon_{p_k}, m^{p_k}_1, m^{p_k}_2>|k=i_1, i_2\}$. We provide the region function $c_r$ and four other region function categories as options.

\textbf{Administrative Region Determination (ARD)}. This task refers to determining which administrative region a coordinate is located in, which involves relevant knowledge of the administrative regions and the ability to associate it with geographical coordinates. For a POI $\bm{p}=<i_p, lat_p, lon_p, c_p, \mathcal{M}_p>$ of the Yelp dataset located in $city_p$, LLMs are asked to answer which city $<lat_p, lon_p>$ is located in. $city_p$ along with other four cities in the same state are provided as options.

\subsection{Spatio-temporal reasoning}

Spatio-temporal reasoning encompasses the ability to understand and reason about the spatial and temporal relationships between entities and events. For example, given a POI and some regions, LLMs should determine which region the POI falls in according to their coordinates and boundary lines. We design four tasks to assess the spatio-temporal reasoning ability of large language models: point-trajectory relationship detection, point-region relationship detection, trajectory-region relationship detection and trajectory identification.

\textbf{Point-Trajectory Relationship Detection (PTRD)}. The task is to determine whether a trajectory passes through a point. To generate a data sample, we downsample the trajectory in the public Xi'an dataset\footnote{https://gaia.didichuxing.com/} into a shorter trajectory $\bm{t}=\{t_1, \cdots, t_n\}$ and construct five points as options. We take $<(lat_i + lat_{i+1})/2, (lon_i+lon_{i+1})/2>$ as the true option, where $<lat_i, lon_i>$ and $<lat_{i+1}, lon_{i+1}>$ are two adjacent points in the trajectory. To construct an error option, we sample a point $t_j=<lat_j, lon_j, time_j>$ from the trajectory and perturb its coordinates with Gaussian noise, \ie, the error option is ${<lat_j+\epsilon_1, lon_j+\epsilon_2>}$, where $\epsilon_1, \epsilon_2 \sim \mathcal{N}(0.01, 0.001)$.

\textbf{Point-Region Relationship Detection (PRRD)}. Given a point and several regions, this task aims to infer which region the point falls in. To generate a data sample, we select $i$ regions $\{\bm{r}_1, \cdots, \bm{r}_i\}$ located in the same city from the EULUC dataset~\citep{gong2020mapping}. Then, a region $\bm{r}_j$ is chosen from these $i$ regions and we randomly selected a point $\bm{p}$ in region $\bm{r}_j$. The coordinates of point $\bm{p}$ and the boundary lines of $i$ regions are used to generate the question texts, and all $i$ regions are provided as options. We construct four sub-datasets by varying the value of $i$ from 2 to 5.

\textbf{Trajectory-Region Relationship Detection (TRRD)}. Given a trajectory and some regions, this task aims to determine which regions the trajectory has passed through chronologically. To construct a data sample, we randomly select five regions $\{\bm{r}_1, \cdots, \bm{r}_5\}$ located in the same city from the EULUC dataset and generate a trajectory $\bm{t}$ by a random walk. The region sequence that $\bm{t}$ passes through and four randomly generated region sequences are provided as options. We construct five sub-datasets by setting the length of $\bm{t}$ to 2, 4, 6, 8 and 10, respectively.

\textbf{Trajectory Identification (TI)}. In this task, we ask LLMs to determine if two point sequences $\bm{t}'$ and $\bm{t}''$ are sampled from the same trajectory. We propose two strategies to construct positive samples (\ie, samples with the answer "Yes") and two strategies to construct negative samples. Specifically, for a trajectory $\bm{t}=<t_1, t_2, \cdots>$ in the Xi'an dataset, we construct two positive samples through downsampling and staggered sampling. For instance, the downsampling strategy use $\bm{t}'=<t_1, t_2, t_3, \cdots>$ and $\bm{t}''=<t_1, t_3, t_5, \cdots>$ to generate the question, while the staggered sampling strategy use $\bm{t}'=<t_1, t_3, t_5, \cdots>$ and $\bm{t}''=<t_2, t_4, t_6, \cdots>$ to generate the question. To construct negative samples, we downsample a trajectory $\bm{t}$ into $\bm{t}'$ and add temporal offsets or spatial offsets to $\bm{t}'$ to obtain $\bm{t}''$.
\subsection{Accurate computation}

In the context of handling spatial-temporal data, accurate computation plays a pivotal role. It focuses on the model's capability to perform precise and complex calculations related to spatial-temporal data. We include two tasks that challenge the model's accuracy in spatial-temporal computations for assessment: direction determination and trajectory-trajectory relationship detection.

\textbf{Direction Determination (DD)}. This task is to determine the direction between two geographical points. To create a data sample, two POIs are randomly chosen from the Yelp dataset, and the model is asked to calculate the corresponding azimuth and to determinate their relative direction based on the calculation result. Eight options are provided for all data samples: north, south, west, east, northeast, northwest, southeast and southwest. 

\textbf{Trajectory-Trajectory Relationship Analysis (TTRA)}. This task is to calculate the number of times two trajectories encounter each other. To construct a data sample, we generate two trajectories $\bm{t}=<t_1, \cdots, t_n>$ and $\bm{t}'=<t_1', \cdots, t_n'>$ through random walks within a certain area. We count it as an encounter if $t_it_{i+1}$ and $t_j't_{j+1}'$ intersect in space and overlap in time, where $1\leq i,j \leq n-1$. We provided the ground truth and other four wrong answers as options.
\subsection{Downstream Applications}

Downstream tasks require the model to not only understand the spatial-temporal context but also apply this understanding to practical applications. We assess this aspect of LLMs through three downstream tasks: trajectory anomaly detection, trajectory classification and trajectory prediction.

\textbf{Trajectory Anomaly Detection (TAD)}. In order to detect anomalous trajectories, LLMs should infer the underlying route and shape from trajectory data. We consider trajectories in Xi'an dataset as normal and perform detours to generate anomalous samples. Specifically, given a trajectory $\bm{t}=<t_1, \cdots, t_n>$, we identify the direction perpendicular to the line connecting $t_1$ and $t_n$, and move the middle one-third of the trajectory along this direction to generate an anomalous sample.

\textbf{Trajectory Classification (TC)}. This task requires the model to comprehensively consider the coordinates, length, speed and other relevant information to distinguish different trajectories. We construct dataset for this task based on the Geolife dataset\footnote{https://www.microsoft.com/en-us/research/publication/geolife-gps-trajectory-dataset-user-guide/}. Due to the input length limitation of LLMs, we downsample each trajectory and ask LLMs to infer what generates the trajectory. Three options are provided: bike, car and pedestrian.

\textbf{Trajectory Prediction (TP)}. This task is to predict the next point based on the historical points of a trajectory involves the ability to model the trajectory patterns and the moving speed. We construct data samples for this task based on the trajectories in the Xi'an dataset. Specifically, we first downsample the each trajectory with a time interval of 30 seconds. Then, for each trajectory $t=<t_1, t_2, \cdots, t_n>$, we ask LLMs to predict the coordinates of $t_j$ according to the historical points $<t_1, \cdots, t_{j-1}>$, where $3 \leq j \leq n$. Note that we do not provide options in this task.
\section{Experiments}
We conduct extensive experiments on \ours to evaluate the spatial-temporal ability of LLMs and to investigate if in-context learning, chain-of-thought and fine-tuning can improve the performance. 

\subsection{Experimental setup}

\textbf{Evaluated models}. We evaluate the performance of two closed-source model, \ie, \textbf{ChatGPT} and \textbf{GPT-4o}, and a set of open-source models: \textbf{Llama-2}~\cite{abs-2307-09288}, \textbf{Vicuna}\footnote{https://lmsys.org/blog/2023-03-30-vicuna/}, \textbf{Gemma}~\cite{abs-2403-08295}, \textbf{Phi-2}, \textbf{ChatGLM2}, \textbf{ChatGLM3}, ~\cite{DuQLDQY022,ZengLDWL0YXZXTM23}, \textbf{Mistral}~\cite{abs-2310-06825}, \textbf{Falcon}~\cite{abs-2311-16867}, \textbf{Deepseek}~\cite{abs-2401-02954}, \textbf{Qwen}~\cite{abs-2309-16609} and \textbf{Yi}~\cite{abs-2403-04652}. More introduction to these models can be found in Appendix B in the supplementary material.


\textbf{Metrics}. We adopt accuracy for tasks other than trajectory prediction. For trajectory prediction, we report absolute error, \ie, the distance in meters between the predicted coordinates and ground truth.

\textbf{Experimental details}. In our experiments, we adopt the precision of FP32 for all LLMs. For all tasks except trajectory prediction, LLMs are expected to answer an option or "Yes"/"No", thus we set the $max\_new\_tokens$ to 15, \ie, the maximum length of the generated new tokens is 15. For trajectory prediction, LLMs should predict the longitude and latitude, and we set the $max\_new\_tokens$ to 50. For other hyperparameters, we adopt the default value of each model. All experiments of open source models are conducted on two NVIDIA H100.
\subsection{Main results}
To investigate the spatio-temporal ability of LLMs, we conduct experiments to evaluate the performance of all models on each task. The main results are shown in Table~\ref{tb:main_result_1} and Table~\ref{tb:main_result_2}.

\begin{table}
  \caption{The performance of \emph{ACC} on knowledge comprehension and spatio-temporal reasoning tasks (bold: best; underline: runner-up). `-' denotes the model failed to answer most questions.}
  \vspace{-1ex}
  \label{tb:main_result_1}
  \centering
  \renewcommand{\arraystretch}{1.1}
  \begin{tabular}{l|c|c|c|c|c|c|c|c}
    \bottomrule
    & \multicolumn{4}{c|}{Knowledge Comprehension} & \multicolumn{4}{c}{Spatio-temporal Reasoning} \\
    \cline{2-9}
    & PCR & PI & URFR & ARD & PTRD & PRRD & TRRD & TI \\
    \hline
    ChatGPT & \underline{0.7926} & 0.5864 & \underline{0.3978} & \underline{0.8358} & \textbf{0.7525} & \textbf{0.9240} & 0.0258 & 0.3342 \\
    GPT-4o & \textbf{0.9588} & \textbf{0.7268} & \textbf{0.6026} & \textbf{0.9656} & - & \underline{0.9188} & 0.1102 & 0.4416\\
    \hline
    ChatGLM2 & 0.2938 & 0.5004 & 0.2661 & 0.2176 & 0.2036 & 0.5216 & \textbf{0.2790} & 0.5000 \\
    ChatGLM3 & 0.4342 & 0.5272 & 0.2704 & 0.2872 & 0.3058 & 0.8244 & 0.1978 & \underline{0.6842} \\
    Phi-2 & - & 0.5267 & - & 0.2988 & - & - & - & 0.5000 \\
    Llama-2-7B & 0.2146 & 0.4790 & 0.2105 & 0.2198 & 0.2802 & 0.6606 & 0.2034 & 0.5486 \\
    Vicuna-7B & 0.3858 & 0.5836 & 0.2063 & 0.2212 & 0.3470 & 0.7080 & 0.1968 & 0.5000\\
    Gemma-2B & 0.2116 & 0.5000 & 0.1989 & 0.1938 & 0.4688 & 0.5744 & 0.2014 & 0.5000 \\
    Gemma-7B & 0.4462 & 0.5000 & 0.2258 & 0.2652 & 0.3782 & 0.9044 & 0.1992 & 0.5000\\
    DeepSeek-7B & 0.2160 & 0.4708 & 0.2071 & 0.1938 & 0.2142 & 0.6424 & 0.1173 & 0.4964 \\
    Falcon-7B & 0.1888 & 0.5112 & 0.1929 & 0.1928 & 0.1918 & 0.4222 & \underline{0.2061} & \textbf{0.7072} \\
    Mistral-7B & 0.3526 & 0.4918 & 0.2168 & 0.3014 & 0.4476 & 0.7098 & 0.0702 & 0.4376\\
    Qwen-7B & 0.2504 & \underline{0.6795} & 0.2569 & 0.2282 & 0.2272 & 0.5762 & 0.1661 & 0.4787 \\
    Yi-6B & 0.3576 & 0.5052 & 0.2149 & 0.1880 & \underline{0.5536} & 0.8264 & 0.1979 & 0.5722\\
    \toprule
  \end{tabular}
  \vspace{-1ex}
\end{table}

\begin{table}
  \caption{The performance of \emph{ACC} and absolute error (in meters) on accurate computation and downstream tasks (bold: best; underline: runner-up). `-' denotes the model failed to directly answer most questions.}
  \vspace{-1ex}
  \label{tb:main_result_2}
  \centering
  \renewcommand{\arraystretch}{1.1}
  \begin{tabular}{l|c|c|c|c|c}
    \bottomrule
    & \multicolumn{2}{c|}{Accurate Computation} & \multicolumn{3}{c}{Downstream Applications} \\
    \hline
    & DD & TTRA & TAD & TC & TP \\
    \hline
    ChatGPT & 0.1698 & 0.1048 & \underline{0.5382} & \textbf{0.4475} & - \\ 
    GPT-4o & \textbf{0.5434} & \textbf{0.3404} & \textbf{0.6016} & - & - \\
    \hline
    ChatGLM2 & 0.1182 & 0.1992 & 0.5000 & 0.3333 & 231.2 \\
    ChatGLM3 & 0.1156 & 0.1828 & 0.5000 & 0.3111 & 224.5 \\
    Phi-2 & 0.1182 & 0.0658 & 0.5000 & 0.3333 & 206.9 \\
    Llama-2-7B & 0.1256 & 0.2062 & 0.5098 & 0.3333 & 189.3 \\
    Vicuna-7B & 0.1106 & 0.1728 & 0.5000 & 0.2558 & 188.1 \\
    Gemma-2B & \underline{0.1972} & 0.2038 & 0.5000 & 0.3333 & 207.7 \\
    Gemma-7B & 0.1182 & 0.1426 & 0.5000 & 0.3333 & \textbf{139.4} \\
    DeepSeek-7B & 0.1972 & 0.1646 & 0.5000 & 0.3333 & 220.8 \\
    Falcon-7B & 0.1365 & 0.2124 & 0.5000 & 0.3309 & 3572.8 \\
    Mistral-7B & 0.1182 & 0.1094 & 0.5000 & 0.3333 & 156.8 \\
    Qwen-7B & 0.1324 & \underline{0.2424} & 0.5049 & \underline{0.3477} & 205.2 \\
    Yi-6B & 0.1284 & 0.2214 & 0.5000 & 0.3333 & \underline{156.2} \\
    \toprule
  \end{tabular}
  \vspace{-2ex}
\end{table}

\textbf{There are significant differences in the performance of different models.} We observe ChatGPT and GPT-4o outperform other models by a large margin on many tasks, \eg, point category recognition, administrative region determination, and point-trajectory relationship detection. There is also a significant difference in performance between open-source models. For instance, Gemma-7B outperforms Qwen-7B on point-region relationship detection with an improvement of 57.0\%, while Qwen-7B outperforms Gemma-7B on point-identification with an improvement of 35.9\%. Although LLMs have the potential to analyze spatio-temporal data, not all models have been adequately trained on relevant corpora and learned corresponding spatio-temporal ability.

\textbf{Model size is important for knowledge and semantic comprehension}. For semantic comprehension, GPT-4o performs better than ChatGPT on all tasks, and ChatGPT outperforms other models on most tasks. The possible reason is that LLMs rely on sufficient parameters to compress and store knowledge, and ChatGPT/GPT-4o has more parameters than other evaluated open-source models. We also observe that Gemma-2B performs poorly on all semantic comprehension tasks, while Gemma-7B, with the same technology but more parameters, achieves higher performance. It supports our conclusion that model size is important for knowledge and semantic comprehension.

\textbf{The evaluated models have difficulty in multi-step reasoning.} The performance of most models on point-region relationship detection is much higher than trajectory-region detection. For instance, the accuracy of ChatGPT is 92.40\% on point-region relationship detection, with only 2.58\% on trajectory-region relationship detection. Note that trajectory-region relationship detection can be achieved by performing point-region relationship detection for each point in the trajectory, thus it is a multi-step reasoning task. The performance on this multi-step task is poor although models such as ChatGPT, GPT-4o, and Gemma-7B can achieve high performance on each step. In conclusion, multi-step spatio-temporal reasoning is difficult for LLMs.

\textbf{Accurate computation and downstream tasks are more challenging}. As shown in Table~\ref{tb:main_result_2}, the accuracy of all models is below 35\% on accurate computation tasks, which is because LLMs are mainly trained on nature language corpus and are not good at computation. Moreover, the performance of evaluated models is also poor on downstream tasks. For instance, the best performance on trajectory anomaly is only 60.16\%, indicating that most evaluated models can not distinguish between normal and anomalous trajectories. The lack of expert knowledge on downstream tasks, \eg, the normal trajectory patterns, leads to their unsatisfactory performance.

\begin{figure}
    \centering
    \includegraphics[width=\textwidth]{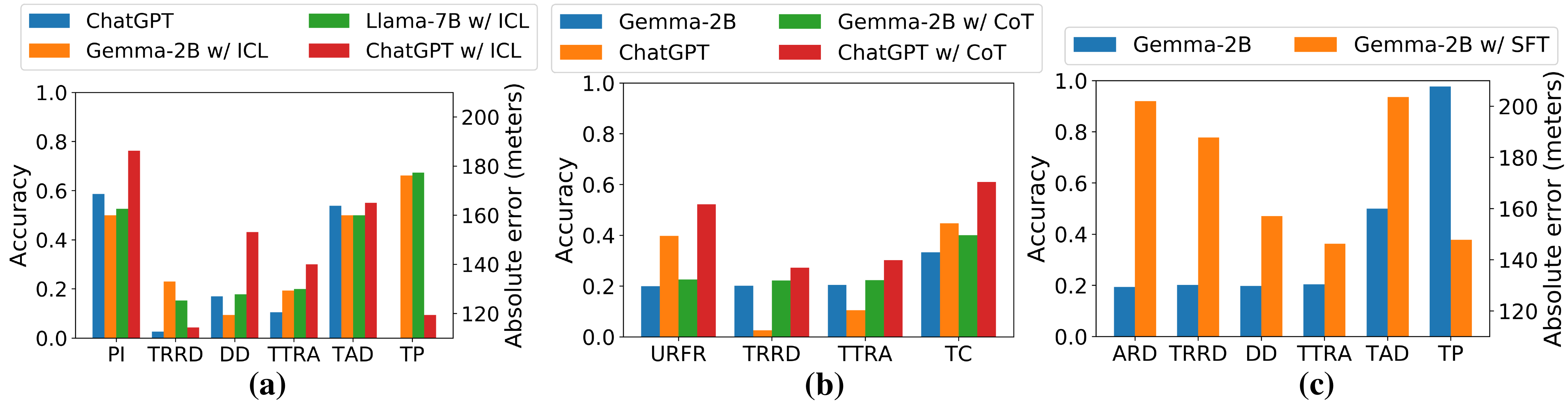}
    \vspace{-3ex}
    \caption{The performance of ACC and absolute error (in meters) in (a) in-context learning evaluation, (b) chain-of-thought evaluation, (c) fine-tuning evaluation.}
    \label{fig:results_1}
    \vspace{-2ex}
\end{figure}

\subsection{In-Context learning evaluation}
Although some evaluated models can perform well on certain tasks, the results in many scenarios are poor. Since LLMs show impressive in-context few-shot learning capacity in previous works, we conduct experiments to investigate if in-context learning can improve the performance of LLMs on \ours. Specifically, we select six tasks where the evaluated models performed poorly and we adopt two-shot prompting. Due to the heavier computation cost caused by the longer context, we only evaluate one closed-source model, ChatGPT, and two open-source models with different model sizes, \ie, Gemma-2B and Llama-2-7B. The results are shown in Fig.~\ref{fig:results_1}(a).

For most tasks, the performance of ChatGPT has been greatly improved with in-context learning. For instance, its performance on POI identification and direction determination has increased from 58.64\% to 76.30\%, and from 16.98\% to 43.16\%, respectively. Moreover, the two-shot prompting also constrains the output, \eg, ChatGPT refuses to answer the questions of trajectory prediction in Table~\ref{tb:main_result_2}, but its absolute error is only 119.4 with two-shot prompting. Although in-context learning is effective for ChatGPT, it is useless on most tasks for Gemma-2B and Llama-2-7B, which is consistent with the phenomenon in previous work that in-context learning is less effective for smaller LLMs~\cite{WeiTBRZBYBZMCHVLDF22}.


\subsection{Chain-of-thought evaluation}
We further conduct experiments to verify if chain-of-thought (CoT) is effective on \ours. Specifically, we evaluate ChatGPT and Gemma-2B with chain-of-thought prompting on several tasks that involve multi-step reasoning: urban region function recognition, trajectory-region relationship detection, trajectory-trajectory relationship analysis and trajectory classification. For each task, we add two samples with a detailed reasoning process in the context, \ie, we implement chain-of-thought by two-shot prompting. For instance, in trajectory classification, we add two samples that contain the reasoning process of calculating the length and average speed of the trajectory. The results are shown in Fig.~\ref{fig:results_1}(b).

We observe the performance of ChatGPT increases significantly in all selected tasks. For instance, its performance with CoT prompting is 52.20\% on urban region function recognition and 61.04\% on trajectory classification, much better than 39.78\% and 44.75\% in Table~\ref{tb:main_result_1} and Table~\ref{tb:main_result_2}. For Gemma-2B, the performance on all selected tasks is also improved. For example, its accuracy increased from 19.89\% to 22.55\% on urban region function recognition and from 33.33\% to 40.05\% on trajectory classification. The results demonstrate the effectiveness of CoT in spatio-temporal analysis.

\subsection{Fine-tuning Evaluation}

While in-context learning and chain-of-thought is less effective for smaller models, we conduct experiments to investigate if fine-tuning can significantly improve the performance on \ours. Specifically, we select several tasks and follow the construction strategies in Section~\ref{sec:construction} to generate 1,2000 samples as the training dataset for each task. We adopt QLoRA~\cite{DettmersPHZ23} to fine-tune the model on the training dataset for each task, with the learning rate of 2e-4, the rank of 8 and NF4 quantization. Due to the very high computational cost and memory usage, we only fine-tune a 2B model for evaluation, \ie, Gemma-2B. The results are shown in Fig.~\ref{fig:results_1}(c).


The performance on all tasks is significantly improved after fine-tuning. For instance, the performance on administrative region determination and direction determination increased from 19.89\% to 91.98\%, and from 19.72\% to 47.08\%, respectively. For trajectory prediction, Gemma-2B achieves the absolute error of 147.8 meters, which is better than all 7B models in Table~\ref{tb:main_result_2}. This confirms LLMs' potential in spatial-temporal analysis and the lack of training on relevant corpora.

\section{Conclusion}
In this work, we propose \ours to assess LLMs' ability in spatio-temporal analysis. \ours consists of 13 tasks and over 60,000 QA pairs, systematically evaluating four dimensions: knowledge comprehension, spatio-temporal reasoning, accurate computation, and downstream applications. We benchmark 13 latest LLMs and the results show their remarkable performance on knowledge comprehension and spatio-temporal reasoning tasks. Our further experiments with in-context learning, chain-of-thought prompting and fine-tuning also prove the great potential of LLMs on other tasks.

\textbf{Limitations}. Due to the rapid evolution of large language models and their enormous computational costs, our assessment is difficult to cover the latest models. For instance, when we started benchmarking, Llama-2 and DeepSeek were the latest models, but now Llama-3 and DeepSeek-V2 have appeared. We have already released our datasets and code, along with the experiments results. We will maintain the project and benchmark more LLMs.

\section*{Acknowledgements}
The data samples of \ours are constructed based on several open-source datasets, \ie, the Yelp Dataset from Yelp Inc., the New Orleans Region Dataset from United States government, the Xi'an Dataset from DiDi Chuxing Inc., the EULUC Dataset from Tsinghua University and the Geolife Dataset from Microsoft Research.

\newpage
\appendix

\section*{Appendix}

\section{Data Format}

\subsection{Prompt template for chatting models}
To make the responses of LLMs controllable and identification of the final answer easier, all data samples in \ours are constructed in the form of text completion. However, there are some chatting models that only support chat completion and do not support text completion, \eg, GPT-4o. For these models, we instruct them to complete the text entered by the human via system prompt. The data samples we constructed are inputted with the role of human, as shown in Table~\ref{tb:chat_completion}.

\begin{table}[h]
  \caption{A prompt template of models that only support chat completion. The {\color{blue}blue} {texts} describe the question. The {\color{brown}brown} texts are the options. The {\color{teal}teal} texts denote the guidance that constrains the output of LLMs.}
  \label{tb:chat_completion}
  \centering
  \begin{tabularx}{\textwidth}{X}
    \bottomrule
    \textbf{System}: "You are a helpful text completion assistant. Please continue writing the text entered by the human."

    \textbf{Human}: "{\color{blue}Question: There is a trajectory, xxxx.} {\color{brown}Options: (1) xxxx, (2) xxxx, (3) xxxx, $\cdots$.} {\color{teal}Please answer one option.} 
    {\color{teal}Answer: The answer is option (}" \cr
    \toprule
  \end{tabularx}
\end{table}

\subsection{Data Examples}

\ours consists of 13 distinct tasks, covering four dimensions: knowledge comprehension, spatio-temporal reasoning, accurate computation and downstream applications. We will provide data samples to illustrate the design of each task.

\subsubsection{Knowledge comprehension}

There are four tasks to assess the knowledge comprehension ability of LLMs in spatio-temporal analysis, \ie, \textbf{Administrative Region Determination (ARD)}, \textbf{POI Category Recognition (PCR)}, \textbf{POI Identification (PI)} and \textbf{Urban Region Function Recognition (URFR)}.

As shown in Table~\ref{sample:ard}, for administrative region determination, we provide the coordinates of a location and ask the model to answer which option the coordinates is located in. The options contain five cities in the same state, which makes this task more challenging. The data sample of POI category recognition is shown in Table~\ref{sample:pcr}. LLMs are asked to predict the category of the POI according to its coordinates and two comments, where each comment contains the comment content and the timestamp. We provide five options and each option is a list of tags such as shopping and skin care. 
In POI identification, we ask the model if two POI are actually the same, where the description of each POI consists of its coordinates and two comments, just as shown in Table~\ref{sample:pi}. A data sample of urban region function recognition is presented in Table~\ref{sample:urfr}, which asks the model to predict the urban region function category according to its boundary lines and the POIs located within it.

\begin{table}
  \caption{A data sample for ARD. The {\color{blue}blue} {texts} describe the question. The {\color{brown}brown} texts are the options. The {\color{teal}teal} texts denote the guidance that constrains the output of LLMs.}
  \label{sample:ard}
  \centering
  \begin{tabularx}{\textwidth}{c|X}
    \bottomrule
    \multirow{11}{*}{Question} & \color{blue}Question: Below is the coordinate location information, and the options of the area where the coordinate may be located:
    
    \color{blue}\{
    
    \color{blue}\quad"latitude": 36.104588,
    
    \color{blue}\quad"longitude": -86.81415,
    
    \color{brown}\quad"options": "(0): Eaton, TN (1): Nashville, TN (2): Sewanee, TN (3): Memphis, TN (4): Knoxville, TN"

    \color{brown}\}
    
    \color{teal}Please answer which area the coordinate is located in. Please just answer the number of your option with no other texts.

    \color{teal}Answer: Option (\cr
    \hline
    \multirow{1}{*}{Answer} &
    1): Nashville, TN \cr
    \toprule
  \end{tabularx}
\end{table}

\begin{table}
  \caption{A data sample for PCR. The {\color{blue}blue} {texts} describe the question. The {\color{brown}brown} texts are the options. The {\color{teal}teal} texts denote the guidance that constrains the output of LLMs.}
  \label{sample:pcr}
  \centering
  \begin{tabularx}{\textwidth}{c|X}
    \bottomrule
    \multirow{28}{*}{Question} & \color{blue}Question: Below is the coordinate location information and related comments of a location, with the options of possible function for this location:

    \color{blue}\{
    
    \color{blue}\quad"latitude": 36.0423589,
    
    \color{blue}\quad"longitude": -86.7788876,
    
    \color{blue}\quad"comment1": \{
    
    \color{blue}\quad\quad"content": "BEST WAX CENTER. ZOIE AND ERIN ARE THE BEST. zoie is calm and friendly makes me feel comfortable all the time erin kills the game with my eyebrows every single time. Everyone asks about my eyebrows thanks to her. Definitely recommend going to Erin and zoie.",
    
    \color{blue}\quad\quad"time": "2021-06-02 00:37:48"
    
    \color{blue}\quad\},
    
    \color{blue}\quad"comment2":\{
    
    \color{blue}\quad\quad"content": "I have done 2 sessions with Erin and LOVE HER! I was so incredibly nervous my first time getting a full bikini wax, but she made me feel so comfortable. She talked through what she was doing, asked me questions, and made the process seem less painful overall. I highly recommend her!!",
    
    \color{blue}\quad\quad"time": "2021-06-18 00:28:02"
    
    \color{blue}\quad\},
    
    \color{brown}\quad"options": "(0): Computers, Shopping, Appliances, Furniture Stores, Home \& Garden (1): Waxing, Hair Removal, Skin Care, Beauty \& Spas (2): Juice Bars \& Smoothies, Food, Vegan, Restaurants, Acai Bowls (3): Discount Store, Shopping, Toy Stores, Food, Candy Stores, Specialty Food (4): Delis, Food, Coffee \& Tea, Sandwiches, Restaurants, Convenience Stores"
    
    \color{brown}\}
    
    \color{teal}Please answer which function the location is. Please just answer the number of your option with no other texts.
    
    Answer: Option ( \cr
    \hline
    \multirow{1}{*}{Answer} & 1): Waxing, Hair Removal, Skin Care, Beauty \& Spas\cr
    \toprule
  \end{tabularx}
\end{table}

\begin{table}
  \caption{A data sample for PI. The {\color{blue}blue} {texts} describe the question. The {\color{brown}brown} texts are the options. The {\color{teal}teal} texts denote the guidance that constrains the output of LLMs.}
  \label{sample:pi}
  \centering
  \begin{tabularx}{\textwidth}{c|X}
    \bottomrule
    \multirow{48}{*}{Question} & \color{blue}Question: Below are two Points of Interest (POI) and related comments.

    \color{blue}POI 1:

    \color{blue}\{

    \color{blue}\quad"latitude": 34.4266787,

    \color{blue}\quad"longitude": -119.7111968,

    \color{blue}\quad"comment1": \{

    \color{blue}\quad\quad"content": "Abby Rappoport helped me achieve a long lost sense of health. I was suffering from debilitating insomnia due to a very stressful job and family requirements. She also was able to get me through a bad bout of bronchitis. She is professional, thorough and clearly seasoned as a healthcare provider. I highly recommend Abby if your situation needs caring attention.",

    \color{blue}\quad\quad"time": "2012-08-09 20:43:27"

    \color{blue}\quad\},
    
    \color{blue}\quad"comment2": \{
    
    \color{blue}\quad\quad"content": "Abby is an amazing practitioner. In a treatment she is really present with me and my concerns. She is caring and thorough. I especially appreciate the exercise, herbs and advice she sends me home with so that my healing can continue outside her office. Abby has helped me with stress related problems and chronic low back pain. Sadly, she moved out of my area but whenever I'm her neck of the woods I take the opportunity to see her.",
    
    \color{blue}\quad\quad"time": "2013-03-01 06:11:05"
    
    \color{blue}\quad\}
    
    \color{blue}\}.
    
    \color{blue}POI2:
    
    \color{blue}\{
    
    \color{blue}\quad"latitude": 34.4266621,
    
    \color{blue}\quad"longitude": -119.711207,
    
    \color{blue}\quad"comment1": \{
    
    \color{blue}\quad\quad"content": "Before buying I looked to see if they had a map off merchants to see where they were located and found no map. If there is one there out is hard to find. I won't buy unless I can tell if members are near me by way of a seeing them onassis map.",
    
    \color{blue}\quad\quad"time": "2014-08-25 00:37:13"
    
    \color{blue}\quad\},
    
    \color{blue}\quad"comment2": \{
    
    \color{blue}\quad\quad"content": "Buyer beware!.... I purchased this card last year and used the buy 1 get 1 free deal and was told it's meant for two people. This was at McConnell's fine ice cream on state street. This guy who's the manager or owner of the business said this deal is meant for you to bring someone along and enjoy the ice cream together and not for you to come in and walk away w/two ice cream cones and pay for one ice cream cone. At the end he said come back w/a friend. He was annoyed.",
    
    \color{blue}\quad\quad"time": "2020-10-09 16:54:26"
    
    \color{blue}\quad\}
    
    \color{blue}\}.
    
    \color{blue}Check whether the two POIs are the same place. Notice that due to the errors, the latitude and longitude may be different although two POI represent the same place.
    
    \color{brown}Please answer "Yes" or "No".
    
    \color{teal}Answer: The answer is " \cr
    \hline
    \multirow{1}{*}{Answer} & No\cr
    \toprule
  \end{tabularx}
\end{table}

\begin{table}[H]
  \caption{A data sample for URFR. The {\color{blue}blue} {texts} describe the question. The {\color{brown}brown} texts are the options. The {\color{teal}teal} texts denote the guidance that constrains the output of LLMs.}
  \label{sample:urfr}
  \centering
  \begin{tabularx}{\textwidth}{c|X}
    \bottomrule
    \multirow{39}{*}{Question} & \color{blue}Question: Below is the coordinate information and related comments of a region, with the options of possible function for this region:

    \color{blue}\{
    
    \color{blue}\quad"region": [(-90.0877900, 29.9689360), (-90.0872427, 29.9689360), (-90.0872427, 29.9696428), (-90.0877900, 29.9696428), (-90.0877900, 29.9689360)]",
    
    \color{blue}\quad"pois": [
    
    \color{blue}\quad\quad\{
    
    \color{blue}\quad\quad\quad"latitude": 29.9694327,
    
    \color{blue}\quad\quad\quad"longitude": -90.0874047,
    
    \color{blue}\quad\quad\quad"comment1": \{
    
    \color{blue}\quad\quad\quad\quad"content": "I cannot day enough about how much I love this place. NOCB popped up on my Instagram feed in 2017 with their Black Friday deals, signed up on a whim and never looked back. The classes are fun and exciting and a great way to get a feel for boxing and each of the trainers here. The gym has become my favorite past time and I love taking my friends in to understand why I'm hooked.",
    
    \color{blue}\quad\quad\quad\quad"time": "2019-12-05 01:43:01"
    
    \color{blue}\quad\quad\quad\},
    
    \color{blue}\quad\quad\quad"comment2": \{
    
    \color{blue}\quad\quad\quad\quad"content": "The best boxing gym in the city! I started boxing a year ago wanting to both get in better shape and learn the skills associated with boxing. I've tried a few places but ultimately settled at NOBC. The positive atmosphere is the first thing you notice about this gym, regardless of if you are a professional or a first time boxer everyone trains together and shares the same passion for boxing, wanting to better themselves through the sport. The gym has everything you need from a weight room, a full boxing ring/equipment, and a cardio/ab area. In a few short weeks training with the owner Chase I have become a better boxer. The gym is clean, friendly, and fun. I plan on training here for years to come.",
    
    \color{blue}\quad\quad\quad\quad"time": "2016-10-29 02:02:17"
    
    \color{blue}\quad\quad\quad\}
    
    \color{blue}\quad\quad\}
    
    \color{blue}\quad],
    
    \color{brown}\quad"options": "(0): Suburban Lake Area Neighborhood Park District (1): Suburban Pedestrian Oriented Corridor Business District (2): Historic Urban Neighborhood Business District (3): Greenway Open Space District (4): Historic Marigny Treme Bywater Commercial District"
    
    \color{brown}\}
    
    \color{teal}Please answer which function the location is. Please just answer the number of your option with no other texts.
    
    Answer: Option (\cr
    \hline
    \multirow{1}{*}{Answer} & 2): Historic Urban Neighborhood Business District\cr
    \toprule
  \end{tabularx}
\end{table}

\subsubsection{Spatio-temporal reasoning}

The dimension of spatio-temporal reasoning consists of four tasks: \textbf{Point-Trajectory Relationship Detection (PTRD)}, \textbf{Point-Region Relationship Detection (PRRD)}, \textbf{Trajectory-Region Relationship Detection (TRRD)} and \textbf{Trajectory Identification (TI)}. The data sample of point-trajectory relationship detection provides a trajectory and five points, then ask the model which point the trajectory passes through, as shown in Table~\ref{sample:ptrd}. A sample of point-region relationship detection is given in Table~\ref{sample:prrd}, which ask the model to determine which region a point falls in according to the boundary lines of the regions and the coordinates of the point. As an enhancement to this task, trajectory-region relationship detection further ask which regions a trajectory passes through chronologically, as shown in Table~\ref{sample:trrd}. 
Trajectory identification aim to determine if two point sequences describe the same trajectory, whose data samples are constructed by four strategies, \ie, downsampling, staggered sampling, spatial offset and temporal offset. By setting different downsampling rate or sampling different points, we can get two point sequences that describe the same trajectory, as shown in Table~\ref{sample:ti_1}. By adding spatial offset or temporal offset to the coordinates or timestamps of the trajectory, we can get another different trajectory, as shown in Table~\ref{sample:ti_2}.

\begin{table}[H]
  \caption{A data sample for PTRD. The {\color{blue}blue} {texts} describe the question. The {\color{brown}brown} texts are the options. The {\color{teal}teal} texts denote the guidance that constrains the output of LLMs.}
  \label{sample:ptrd}
  \centering
  \begin{tabularx}{\textwidth}{c|X}
    \bottomrule
    \multirow{12}{*}{Question} & \color{blue}Question: \color{blue}The following is a sequence of points sampled from a trajectory and the meaning of each point is (longitude, latitude, timestamp): [(108.91226, 34.25924, 1477967031), (108.92136, 34.25929, 1477967109), (108.92268, 34.26271, 1477967184), (108.92247, 34.27329, 1477967274), (108.92732, 34.27659, 1477967352), (108.93702, 34.27663, 1477967430), (108.9435, 34.27682, 1477967505), (108.95271, 34.27686, 1477967586), (108.95937, 34.27675, 1477967662), (108.97203, 34.27726, 1477967767)]. 
    
    \color{brown}The trajectory passes through one of the following points: (1) Point 1 (108.93244, 34.28307); (2) Point 2 (108.95336, 34.28628); (3) Point 3 (108.93661, 34.28624); (4) Point 4 (108.91681, 34.259265); (5) Point 5 (108.92387, 34.26896);
    
    \color{blue}Please answer which option the trajectory passes through. 
    
    \color{teal}Answer: The trajectory passes through Point \cr
    \hline
    \multirow{1}{*}{Answer} & 4 \cr
    \toprule
  \end{tabularx}
\end{table}

\begin{table}[H]
  \caption{A data sample for PRRD. The {\color{blue}blue} {texts} describe the question. The {\color{teal}teal} texts denote the guidance that constrains the output of LLMs.}
  \label{sample:prrd}
  \centering
  \begin{tabularx}{\textwidth}{c|X}
    \bottomrule
    \multirow{10}{*}{Question} & \color{blue}Question: There are several regions, and the boundary lines of each region are presented in the form of a list of (longitude, latitude) below:
    
    \color{blue}Region 1: [(104.2483, 33.2447), (104.2481, 33.2440), (104.2470, 33.2438), (104.2466, 33.2440), (104.2464, 33.2443), (104.2463, 33.2446), (104.2477, 33.2456)]
    
    \color{blue}Region 2: [(104.2446, 33.2471), (104.2453, 33.2460), (104.2456, 33.2450), (104.2451, 33.2451), (104.2448, 33.2454), (104.2443, 33.2457), (104.2437, 33.2459), (104.2432, 33.2462), (104.2431, 33.2465)]
    
    \color{blue}Now there is a point with longitude 104.2444 and latitude 33.2460. Please directly answer the number of the region that this point falls in. 
    
    \color{teal}Answer: The point falls in Region \cr
    \hline
    \multirow{1}{*}{Answer} & 2\cr
    \toprule
  \end{tabularx}
\end{table}

\begin{table}[H]
  \caption{A data sample for TRRD. The {\color{blue}blue} {texts} describe the question. The {\color{brown}brown} texts are the options. The {\color{teal}teal} texts denote the guidance that constrains the output of LLMs.}
  \label{sample:trrd}
  \centering
  \begin{tabularx}{\textwidth}{c|X}
    \bottomrule
    \multirow{25}{*}{Question} & \color{blue}Question: There are several regions, and the boundary lines of each region are presented in the form of a list of (longitude, latitude) below:
    
    \color{blue}Region 1: [(104.2483209, 33.2446592), (104.2480514, 33.2440436), (104.2469734, 33.2437741), (104.2465616, 33.2440436), (104.2464345, 33.2443130), (104.2462657, 33.2445689), (104.2477476, 33.2456337)]
    
    \color{blue}Region 2: [(104.2446473, 33.2470611), (104.2452599, 33.2459617), (104.2456260, 33.2449552), (104.2450870, 33.2451215), (104.2448175, 33.2453910), (104.2442785, 33.2456605), (104.2437395, 33.2459300), (104.2432005, 33.2461995), (104.2430696, 33.2464690)]
    
    \color{blue}Region 3: [(104.2476758, 33.2457578), (104.2459598, 33.2450870), (104.2454075, 33.2460224), (104.2447964, 33.2471191), (104.2465063, 33.2478088), (104.2476758, 33.2457578)]
    
    \color{blue}Region 4: [(104.2445777, 33.2471861), (104.2427577, 33.2466877), (104.2423098, 33.2477300), (104.2424400, 33.2481779), (104.2433484, 33.2491652), (104.2433517, 33.2491689), (104.2436447, 33.2488824), (104.2442290, 33.2478118)]
    
    \color{blue}Region 5: [(104.2464353, 33.2479333), (104.2447267, 33.2472441), (104.2443780, 33.2478698), (104.2438019, 33.2489228), (104.2436336, 33.2494729), (104.2451994, 33.2499855), (104.2458120, 33.2490264), (104.2464353, 33.2479333)]
    
    \color{blue}Now there is a trajectory presented in the form of a list of (longitude, latitude): [(104.2453154, 33.2468798), (104.2431636, 33.2476642), (104.2448701, 33.2483024), (104.2427480, 33.2486476), (104.2466176, 33.2489308)]. Note that although we only provide the coordinates of some discrete points, the trajectory is actually continuous. 
    
    \color{brown}Please answer which regions it has passed through in chronological order: (1) [3, 2, 1, 5], (2) [3, 4], (3) [3, 4, 2, 3], (4) [3, 2, 4, 2, 1], (5) [3, 4, 5]. 
    
    \color{teal}Answer only one option with no other texts. Answer: Option (\cr
    \hline
    \multirow{1}{*}{Answer} & 5): [3, 4, 5]\cr
    \toprule
  \end{tabularx}
\end{table}

\begin{table}[H]
  \caption{Data samples constructed by downsampling and staggered sampling for TI. The {\color{blue}blue} {texts} describe the question. The {\color{brown}brown} texts are the options. The {\color{teal}teal} texts denote the guidance that constrains the output of LLMs.}
  \label{sample:ti_1}
  \centering
  \begin{tabularx}{\textwidth}{c|X}
    \bottomrule
    \multicolumn{2}{l}{Downsampling} \cr
    \hline
    \multirow{23}{*}{Question} & \color{blue}Question: There are two point sequences and each sequence is sampled from a trajectory. The meaning of each point is (longitude, latitude, time stamp). Please answer whether these two sequences are sampled from the same trajectory. 
    
    \color{blue}Sequence 1: [(108.91226, 34.25924, 1477967031), (108.92136, 34.25929, 1477967106), (108.92277, 34.26197, 1477967178), (108.92248, 34.27254, 1477967265), (108.92586, 34.27659, 1477967340), (108.93587, 34.27662, 1477967415), (108.94108, 34.27671, 1477967487), (108.95088, 34.27682, 1477967564), (108.95635, 34.27691, 1477967638)], 
    
    \color{blue}Sequence 2: [(108.91226, 34.25924, 1477967031), (108.91715, 34.25925, 1477967067), (108.92136, 34.25929, 1477967106), (108.92275, 34.25931, 1477967142), (108.92277, 34.26197, 1477967178), (108.92257, 34.2661, 1477967217), (108.92248, 34.27254, 1477967265), (108.92307, 34.27581, 1477967301), (108.92586, 34.27659, 1477967340), (108.93109, 34.2766, 1477967379), (108.93587, 34.27662, 1477967415), (108.93958, 34.27668, 1477967451), (108.94108, 34.27671, 1477967487), (108.94591, 34.27739, 1477967523), (108.95088, 34.27682, 1477967564), (108.95329, 34.27687, 1477967602), (108.95635, 34.27691, 1477967638)]. 
    
    \color{blue}You can confirm if their routes are the same by checking if sequence 1 passes through each point in sequence 2. Then, check if their timestamps are consistent. Finally, answer whether they are sampled from the same trajectory. 
    
    \color{brown}Please answer "Yes" or "No".
    
    \color{teal}Answer: The answer is "\cr
    \hline
    \multirow{1}{*}{Answer} & Yes\cr
    \toprule
    \multicolumn{2}{l}{Staggered Sampling} \cr
    \hline
    \multirow{23}{*}{Question} & \color{blue}Question: There are two point sequences and each sequence is sampled from a trajectory. The meaning of each point is (longitude, latitude, time stamp). Please answer whether these two sequences are sampled from the same trajectory. 
    
    \color{blue}Sequence 1: [(108.91267, 34.25924, 1477967034), (108.91758, 34.25925, 1477967070), (108.92136, 34.25929, 1477967109), (108.923, 34.25931, 1477967145), (108.92273, 34.26228, 1477967181), (108.92256, 34.26648, 1477967220), (108.92248, 34.27273, 1477967268), (108.92317, 34.27594, 1477967304), (108.92621, 34.27659, 1477967343), (108.93137, 34.27661, 1477967382), (108.93614, 34.27663, 1477967418), (108.93984, 34.27668, 1477967454), (108.94133, 34.27671, 1477967490), (108.94635, 34.27738, 1477967526), (108.95162, 34.27684, 1477967568), (108.95344, 34.27687, 1477967605), (108.95665, 34.27689, 1477967641)],
    
    \color{blue}Sequence 2: [(108.91226, 34.25924, 1477967031), (108.91715, 34.25925, 1477967067), (108.92136, 34.25929, 1477967106), (108.92275, 34.25931, 1477967142), (108.92277, 34.26197, 1477967178), (108.92257, 34.2661, 1477967217), (108.92248, 34.27254, 1477967265), (108.92307, 34.27581, 1477967301), (108.92586, 34.27659, 1477967340), (108.93109, 34.2766, 1477967379), (108.93587, 34.27662, 1477967415), (108.93958, 34.27668, 1477967451), (108.94108, 34.27671, 1477967487), (108.94591, 34.27739, 1477967523), (108.95088, 34.27682, 1477967564), (108.95329, 34.27687, 1477967602), (108.95635, 34.27691, 1477967638)]. 
    
    \color{blue}You can confirm if their routes are the same by checking if sequence 1 passes through each point in sequence 2. Then, check if their timestamps are consistent. Finally, answer whether they are sampled from the same trajectory. 
    
    \color{brown}Please answer "Yes" or "No". 
    
    \color{teal}Answer: The answer is " \cr
    \hline
    \multirow{1}{*}{Answer} & Yes \cr
    \toprule
  \end{tabularx}
\end{table}

\begin{table}[H]
  \caption{Data samples constructed through spatial or temporal offset for TI. The {\color{blue}blue} {texts} describe the question. The {\color{brown}brown} texts are the options. The {\color{teal}teal} texts denote the guidance that constrains the output of LLMs.}
  \label{sample:ti_2}
  \centering
  \begin{tabularx}{\textwidth}{c|X}
    \bottomrule
    \multicolumn{2}{l}{Spatial Offset} \cr
    \hline
    \multirow{23}{*}{Question} & \color{blue}Question: There are two point sequences and each sequence is sampled from a trajectory. The meaning of each point is (longitude, latitude, time stamp). Please answer whether these two sequences are sampled from the same trajectory. 
    
    Sequence 1: [(108.91226, 34.25924, 1477967031), (108.91715, 34.25925, 1477967067), (108.92136, 34.25929, 1477967106), (108.92275, 34.25931, 1477967142), (108.92277, 34.26197, 1477967178), (108.92257, 34.2661, 1477967217), (108.94056, 34.28908, 1477967265), (108.94115, 34.29235, 1477967301), (108.94394, 34.29313, 1477967340), (108.94917, 34.29314, 1477967379), (108.95395, 34.29316, 1477967415), (108.95766, 34.29322, 1477967451), (108.95916, 34.29325, 1477967487), (108.96399, 34.29393, 1477967523), (108.96896, 34.29336, 1477967564), (108.97137, 34.29341, 1477967602), (108.95635, 34.27691, 1477967638)], 
    
    Sequence 2: [(108.91226, 34.25924, 1477967031), (108.91715, 34.25925, 1477967067), (108.92136, 34.25929, 1477967106), (108.92275, 34.25931, 1477967142), (108.92277, 34.26197, 1477967178), (108.92257, 34.2661, 1477967217), (108.92248, 34.27254, 1477967265), (108.92307, 34.27581, 1477967301), (108.92586, 34.27659, 1477967340), (108.93109, 34.2766, 1477967379), (108.93587, 34.27662, 1477967415), (108.93958, 34.27668, 1477967451), (108.94108, 34.27671, 1477967487), (108.94591, 34.27739, 1477967523), (108.95088, 34.27682, 1477967564), (108.95329, 34.27687, 1477967602), (108.95635, 34.27691, 1477967638)]. 
    
    You can confirm if their routes are the same by checking if sequence 1 passes through each point in sequence 2. Then, check if their timestamps are consistent. Finally, answer whether they are sampled from the same trajectory. 
    
     \color{brown}Please answer "Yes" or "No".
    
     \color{teal}Answer: The answer is "\cr
    \hline
    \multirow{1}{*}{Answer} & No\cr
    \toprule
    \multicolumn{2}{l}{Temporal Offset} \cr
    \hline
    \multirow{23}{*}{Question} & \color{blue}Question: There are two point sequences and each sequence is sampled from a trajectory. The meaning of each point is (longitude, latitude, time stamp). Please answer whether these two sequences are sampled from the same trajectory. 
    
    Sequence 1: [(108.91226, 34.25924, 1477967031), (108.91715, 34.25925, 1477967067), (108.92136, 34.25929, 1477967106), (108.92275, 34.25931, 1477967142), (108.92277, 34.26197, 1477967178), (108.92257, 34.2661, 1477967217), (108.92248, 34.27254, 1477967265), (108.92307, 34.27581, 1477967301), (108.92586, 34.27659, 1477967340), (108.93109, 34.2766, 1477967379)], 
    
    Sequence 2: [(108.91226, 34.25924, 1478006153), (108.91715, 34.25925, 1478006189), (108.92136, 34.25929, 1478006228), (108.92275, 34.25931, 1478006264), (108.92277, 34.26197, 1478006300), (108.92257, 34.2661, 1478006339), (108.92248, 34.27254, 1478006387), (108.92307, 34.27581, 1478006423), (108.92586, 34.27659, 1478006462), (108.93109, 34.2766, 1478006501)]. 
    
    You can confirm if their routes are the same by checking if sequence 1 passes through each point in sequence 2. Then, check if their timestamps are consistent. Finally, answer whether they are sampled from the same trajectory. 
    
     \color{brown}Please answer "Yes" or "No".
    
     \color{teal}Answer: The answer is "\cr
    \hline
    \multirow{1}{*}{Answer} & No \cr
    \toprule
  \end{tabularx}
\end{table}

\subsubsection{Accurate computation}

The assessing of accurate computation involves two tasks: \textbf{Direction Determination (DD)} and \textbf{Trajectory-Trajectory Relationship Analaysis (TTRA)}. Direction determination aim to predict the relative direction between two given coordinates, as shown in Table~\ref{sample:dd}. For trajectory-trajectory relationship analysis, two trajectories are given and the model is asked to count how many times they intersect, as shown in Table~\ref{sample:ttra}.

\begin{table}
  \caption{A data sample for DD. The {\color{blue}blue} {texts} describe the question. The {\color{brown}brown} texts are the options. The {\color{teal}teal} texts denote the guidance that constrains the output of LLMs.}
  \label{sample:dd}
  \centering
  \begin{tabularx}{\textwidth}{c|X}
    \bottomrule
    \multirow{7}{*}{Question} & \color{blue}Question: A has a longitude of 115.6249 and a latitude of 33.1811, while B has a longitude of 114.3897 and a latitude of 36.085839. Therefore, B is in the () from A. Please choose the correct answer from the following options and fill it in parentheses. 
    
    \color{brown}(1) North, (2) Northeast, (3) East, (4) Southeast, (5) South, (6) Southwest, (7) West, (8) Northwest. 
    
    \color{teal}Please directly give me the number of your option with no other texts. 
    
    \color{teal}Answer: Option (\cr
    \hline
    \multirow{1}{*}{Answer} & 1) North \cr
    \toprule
  \end{tabularx}
\end{table}

\begin{table}
  \caption{A data sample for TTRA. The {\color{blue}blue} {texts} describe the question. The {\color{brown}brown} texts are the options. The {\color{teal}teal} texts denote the guidance that constrains the output of LLMs.}
  \label{sample:ttra}
  \centering
  \begin{tabularx}{\textwidth}{c|X}
    \bottomrule
    \multirow{14}{*}{Question} & \color{blue}Question: There are two trajectories presented in the form of a list of (longitude, latitude, timestamp) below:
    
    \color{blue}trajectory A: [(104.24490, 33.24652, 1683618155), (104.24440, 33.24504, 1683619121), (104.24420, 33.24477, 1683620129), (104.24600, 33.24515, 1683621109), (104.24667, 33.24498, 1683622143)]
    
    \color{blue}trajectory B: [(104.24458, 33.24707, 1683618164), (104.24242, 33.24675, 1683619137), (104.24375, 33.24676, 1683620199), (104.24522, 33.24833, 1683621179), (104.24615, 33.24663, 1683622182)]
    
    \color{blue}Please calculate the number of times these two trajectories intersect, and choose your answer from following options:
    
    \color{brown}(1) 2 times, (2) 3 times, (3) 4 times, (4) 0 times, (5) 1 times.
    
    \color{teal}Note that two trajectories intersect if and only if they pass through the same point at the same timestamp. Give me your option with no other texts. 
    
    \color{teal}Answer: Option (\cr
    \hline
    \multirow{1}{*}{Answer} & 4) 0 times \cr
    \toprule
  \end{tabularx}
\end{table}

\subsubsection{Downstream Applications}

We select three downstream applications for evaluation: \textbf{Trajectory Anomaly Detection (TAD)}, \textbf{Trajectory Classification (TC)} and \textbf{Trajectory Prediction (TP)}. As shown in Table~\ref{sample:tad} and Table~\ref{sample:tc}, given a trajectory, trajectory anomaly detection and trajectory classification aims to infer if the trajectory is anomalous and the source of the trajectory, respectively. For trajectory prediction, the model is asked to predict the next point of a trajectory according to the historical points, as shown in Table~\ref{sample:tp}.

\begin{table}[H]
  \caption{A data sample for TAD. The {\color{blue}blue} {texts} describe the question. The {\color{brown}brown} texts are the options. The {\color{teal}teal} texts denote the guidance that constrains the output of LLMs.}
  \label{sample:tad}
  \centering
  \begin{tabularx}{\textwidth}{c|X}
    \bottomrule
    \multirow{14}{*}{Question} & \color{blue}Question: Below is a trajectory generated by a taxi, and each point in this trajectory is a tuple of (longitude, latitude, timestamp):
    
    [(108.91226, 34.25924, 1477967031), (108.91715, 34.25925, 1477967067), (108.92136, 34.25929, 1477967106), (108.92275, 34.25931, 1477967142), (108.92277, 34.26197, 1477967178), (108.92257, 34.2661, 1477967217), (108.92248, 34.27254, 1477967265), (108.92307, 34.27581, 1477967301), (108.92586, 34.27659, 1477967340), (108.93109, 34.2766, 1477967379), (108.93587, 34.27662, 1477967415), (108.93958, 34.27668, 1477967451), (108.94108, 34.27671, 1477967487), (108.94591, 34.27739, 1477967523), (108.95088, 34.27682, 1477967564), (108.95329, 34.27687, 1477967602), (108.95635, 34.27691, 1477967638), (108.96059, 34.27669, 1477967677), (108.96856, 34.277, 1477967737), (108.97323, 34.27733, 1477967776), (108.97674, 34.27742, 1477967812), (108.97917, 34.27868, 1477967854)].
    
    The trajectory is anomalous if there is a detour, otherwise the trajectory is normal. Please answer if this trajectory is anomalous or normal. 
    
    \color{brown}Please answer \"This trajectory is normal\" or \"This trajectory is anomalous\" with no other texts.
    
    \color{teal}Answer: This trajectory is\cr
    \hline
    \multirow{1}{*}{Answer} & normal \cr
    \toprule
  \end{tabularx}
\end{table}

\begin{table}[H]
  \caption{A data sample for TC. The {\color{blue}blue} {texts} describe the question. The {\color{brown}brown} texts are the options. The {\color{teal}teal} texts denote the guidance that constrains the output of LLMs.}
  \label{sample:tc}
  \centering
  \begin{tabularx}{\textwidth}{c|X}
    \bottomrule
    \multirow{18}{*}{Question} & \color{blue}Question: The following is a sequence of points sampled from a trajectory, and the meaning of each point is (longitude, latitude, timestamp):
    
    [(116.3324016, 40.0743183, 1225573207), (116.3324566, 40.0743099, 1225573208), (116.3326216, 40.0742966, 1225573212), (116.3328333, 40.0742683, 1225573216), (116.3330533, 40.0742516, 1225573220), (116.3332683, 40.0742699, 1225573224), (116.3334999, 40.0742583, 1225573228), (116.3337183, 40.0742583, 1225573232), (116.3339750, 40.0742033, 1225573236), (116.3341916, 40.0742249, 1225573240), (116.3343866, 40.0742749, 1225573244), (116.3345883, 40.0743099, 1225573248), (116.3347933, 40.0742966, 1225573252), (116.3350016, 40.0743049, 1225573256), (116.3352266, 40.0743299, 1225573260), (116.3354566, 40.0743183, 1225573264), (116.3356466, 40.0743099, 1225573268), (116.3358816, 40.0743150, 1225573272)].
    
    \color{brown}The trajectory is generated by one of the following option: (1) car, (2) bike, (3) pedestrian.
    
    \color{teal}Please calculate the length and the average speed of the trajectory, and answer which option is most likely to generate this trajectory.
    
    Answer: The trajectory is most likely to be generated by Option (\cr
    \hline
    \multirow{1}{*}{Answer} & 2 \cr
    \toprule
  \end{tabularx}
\end{table}

\begin{table}[H]
  \caption{A data sample for TP. The {\color{blue}blue} {texts} describe the question. The {\color{brown}brown} texts are the options. The {\color{teal}teal} texts denote the guidance that constrains the output of LLMs.}
  \label{sample:tp}
  \centering
  \begin{tabularx}{\textwidth}{c|X}
    \bottomrule
    \multirow{6}{*}{Question} & \color{blue}Question: Below is an ongoing trajectory generated by a taxi, and each point in this trajectory is a tuple of (longitude, latitude, timestamp):
    
    [(108.92788, 34.23136, 1477956224), (108.92637, 34.23206, 1477956254), (108.92599, 34.23226, 1477956284), (108.92527, 34.23263, 1477956314)].
    
    Please predict the longitude and latitude of the next point.
    
    \color{teal}Answer: The longitude and latitude of the next point is\cr
    \hline
    \multirow{1}{*}{Answer} & [108.92384, 34.23327] \cr
    \toprule
  \end{tabularx}
\end{table}
\section{Experimental Details}

\subsection{Evaluated models}
We evaluate two closed-source models and a set of open-source models. The two closed-source models are \textbf{ChatGPT} (gpt-3.5-turbo-1106) and \textbf{GPT-4o} (gpt-4o-2024-05-13), both developed by OpenAI. For open-source models, we first select two models from the popular Llama family, \ie, \textbf{Llama-2-7B} and \textbf{Vicuna-7B}, which are released by Meta and Large Model Systems Organization, respectively. Then, we include \textbf{Gemma-2B} and \textbf{Gemma-7B},  which are developed by Google DeepMind, based on Gemini research and technology. \textbf{Phi-2}, a model with only 2.7 billion parameters proposed by Microsoft Research, is evaluated to investigate the performance of lightweight language models. We also evaluate \textbf{ChatGLM2} and \textbf{ChatGLM3}, two open bilingual language models with 6B parameters based on General Language Model (GLM). Moreover, \textbf{Mistral-7B}, a large language model developed by Mistral AI, is also included. Futhermore, other baselines includes textbf{Falcon-7B}, a LLM developed by Technology Innovation Institute; \textbf{Deepseek-7B}, the language model presented by Deepseek AI; \textbf{Qwen-7B}, the language model of Alibaba and \textbf{Yi-6B}, an open foundation model by 01.AI. All experiments about the open-source models are conducted on modelscope~\footnote{https://github.com/modelscope/modelscope}. The details about the downloading of all open-source models, the code for reproducing our experiments, and the benchmark datasets can be found at \url{https://github.com/LwbXc/STBench}.

\subsection{Detailed results}
There are some tasks that consists of several sub-datasets. Specifically, for the point-region relationship detection task, we vary the number of regions from 2 to 5 to obtain 4 sub-datasets. In the trajectory-region relationship detection task, the trajectory length is set to 2, 4, 6, 8, 10 to construct five sub-datasets. Moreover, we adopt four strategies to construct the data samples for trajectory identification, resulting in four sub-datasets.

\begin{table}
  \caption{The performance of \emph{ACC} on sub-datasets of point-region relationship detection and trajectory-region relationship detection. $r$ denotes the number of regions and $l$ denotes the length of the trajectory.}
  \label{tb:prrd_trrd_basic}
  \centering
  \renewcommand{\arraystretch}{1.1}
  \resizebox{\textwidth}{35mm}{
  \begin{tabular}{l|c|c|c|c|c|c|c|c|c}
    \bottomrule
    & \multicolumn{4}{c|}{PRRD} & \multicolumn{5}{c}{TRRD} \\
    \cline{2-10}
    & $r=2$ & $r=3$ & $r=4$ & $r=5$ & $l=2$ & $l=4$ & $l=6$ & $l=8$ & $l=10$ \\
    \hline
    ChatGPT & 0.9568 & 0.9176 & 0.8864 & 0.9352 & 0.0536 & 0.0312 & 0.0136 & 0.0160 & 0.0144\\
    GPT-4o & 0.9224 & 0.9160 & 0.9096 & 0.9272 & 0.2504 & 0.1088 & 0.0680 & 0.0624 & 0.0616\\
    \hline
    ChatGLM2 & 0.5624 & 0.6144 & 0.4216 & 0.4880 & 0.3104 & 0.2736 & 0.2536 & 0.2880 & 0.2696 \\
    ChatGLM3 & 0.9096 & 0.8400 & 0.7328 & 0.8152 & 0.2256 & 0.2144 & 0.2032 & 0.1784 & 0.1672 \\
    Phi-2 & - & - & - & - & - & - & - & - & -\\
    Llama-2-7B & 0.5888 & 0.6504 & 0.6208 & 0.7824 & 0.2128 & 0.2088 & 0.1936 & 0.2072 & 0.1944 \\
    Vicuna-7B & 0.7840 & 0.7160 & 0.5920 & 0.7400 & 0.1864 & 0.2032 & 0.1832 & 0.2008 & 0.2104\\
    Gemma-2B & 0.7024 & 0.5696 & 0.5408 & 0.4848 & 0.2096 & 0.1904 & 0.2168 & 0.1960 & 0.1944 \\
    Gemma-7B & 0.9056 & 0.9072 & 0.8904 & 0.9144 & 0.2096 & 0.1856 & 0.2128 & 0.1952 & 0.1928 \\
    DeepSeek-7B & 0.8544 & 0.5968 & 0.5184 & 0.6000 & 0.1504 & 0.1328 & 0.1001 & 0.1088 & 0.0944\\
    Falcon-7B & 0.5602 & 0.4344 & 0.3296 & 0.3647 & 0.1995 & 0.2110 & 0.2090 & 0.2062 & 0.2046 \\
    Mistral-7B & 0.5336 & 0.7104 & 0.7256 & 0.8696 & 0.1896 & 0.0704 & 0.0320 & 0.0304 & 0.0288 \\
    Qwen-7B & 0.6448 & 0.5752 & 0.5184 & 0.5662 & 0.2544 & 0.1544 & 0.1312 & 0.1496 & 0.1408 \\
    Yi-6B & 0.9192 & 0.8008 & 0.7560 & 0.8296 & 0.2184 & 0.1816 & 0.1744 & 0.1672 & 0.1816\\
    \toprule
  \end{tabular}
  }
\end{table}

\subsubsection{Basic prompt}
The results on these sub-datasets with basic prompt are shown in Table~\ref{tb:prrd_trrd_basic} and Table~\ref{tb:ti_basic}. For point-region relationship detection, we observe that most models achieve higher performance on sub-datasets with fewer regions, which is in line with our intuition. But there are also exceptions, \eg, Mistral-7B achieve higher performance with more regions. For the trajectory-region relationship detection, the performance of most models decreases with larger trajectory length, since longer trajectory makes the task more challenging. For trajectory identification, we observe that some models consistently answer "Yes" or "No", regardless of the question, \eg, ChatGLM2 and Phi-2. We also observe that different models have different characteristics. For instance, GPT-4o can find out spatial offset in trajectories, but it failed to identify the temporal offset. ChatGLM3 is good at identifying downsampling, staggered sampling and temporal offset, but it did not recognize the spatial offset. No evaluated model can achieve high performance on all four sub-datasets.

\begin{table}
  \caption{The performance of \emph{ACC} on sub-datasets of trajectory identification.}
  \label{tb:ti_basic}
  \centering
  \renewcommand{\arraystretch}{1.1}
  {
  \begin{tabular}{l|c|c|c|c}
    \bottomrule
    & Downsampling & Staggered & Temporal& Spatial\\
    \hline
    ChatGPT & 0.1784 & 0.0016 & 0.8464 & 0.3104 \\
    GPT-4o & 0.1624 & 0.5840 & 0.0280 & 0.9920 \\
    \hline
    ChatGLM2 & 0.0000 & 0.0000 & 1.0000 & 1.0000 \\
    ChatGLM3 & 0.9992 & 0.9368 & 0.8008 & 0.0000 \\
    Phi-2 & 1.0000 & 1.0000 & 0.0000 & 0.0000\\
    Llama-2-7B & 0.1952 & 0.9992 & 1.0000 & 0.0000\\
    Vicuna-7B & 0.0000 & 0.0000 & 1.0000 & 1.0000 \\
    Gemma-2B & 0.0000 & 0.0000 & 1.0000 & 1.0000 \\
    Gemma-7B & 1.0000 & 1.0000 & 0.0000 & 0.0000 \\
    DeepSeek-7B & 1.0000 & 0.9856 & 0.0000 & 0.0000\\
    Falcon-7B & 0.8264 & 0.0024 & 1.0000 & 1.0000\\
    Mistral-7B & 0.0056 & 0.0000 & 1.0000 & 0.7448\\
    Qwen-7B & 0.3992 & 0.3395 & 0.6047 & 0.5714\\
    Yi-6B & 0.9888 & 0.8856 & 0.0000 & 0.4144\\
    \toprule
  \end{tabular}
  }
\end{table}

\subsubsection{In-context learning}
The results on sub-datasets of trajectory-region relationship detection with in-context learning are shown in Table~\ref{tb:trrd_icl}. We find that in-context learning significantly improve the performance of ChatGPT on sub-datasets with the trajectory length of 2, but it is useless for sub-datasets with longer trajectories. We also observe that in-context learing slightly improve the performance of Gemma-2B on sub-datasets with trajectory length larger than 2, which is exactly opposite to ChatGPT.

\begin{table}
  \caption{The performance of \emph{ACC} on sub-datasets of trajectory-region relationship detection with in-context learning, chain-of-thought prompting and fine-tuning. $l$ denotes the length of the trajectory.}
  \label{tb:trrd_icl}
  \centering
  \renewcommand{\arraystretch}{1.1}
  \begin{tabular}{l|c|c|c|c|c}
    \bottomrule
    & $l=2$ & $l=4$ & $l=6$ & $l=8$ & $l=10$ \\
    \hline
    ChatGPT w/ ICL& 0.1432 & 0.0408 & 0.0120 & 0.0080 & 0.0088\\
    Llama-2-7B w/ ICL & 0.2000 & 0.1688 & 0.1328 & 0.1232 & 0.1376 \\
    Gemma-2B w/ ICL & 0.2088 & 0.2472 & 0.2376 & 0.2384 & 0.2200 \\
    \hline
    ChatGPT w/ CoT & 0.7504 & 0.2520 & 0.1584 & 0.1112 & 0.0872\\
    Gemma-2B w/ CoT & 0.2210 & 0.2564 & 0.2287 & 0.1910 & 0.2125 \\
    \hline
    Gemma-2B w/ SFT & 0.7560 & 0.8104 & 0.8072 & 0.7512 & 0.7640 \\
    \toprule
  \end{tabular}
\end{table}

\subsubsection{Chain-of-thought}
The results on sub-datasets of trajectory-region relationship detection with chain-of-though prompting are shown in Table~\ref{tb:trrd_icl}. We observe that CoT further significantly boost the performance of ChatGPT on most sub-datasets. With the trajectory length increases, the performance of ChatGPT with CoT decreases sharply. For Gemma-2B, CoT does not further improve its performance compared with ICL.

\subsubsection{Fine-tuning}
The results on sub-datasets of trajectory-region relationship detection after fine-tuning are shown in Table~\ref{tb:trrd_icl}. We observe fine-tuning significantly improve the performace of Gemma-2B on all sub-datasets. The performance after fine-tuning does not decreases with larger trajectory length.

\end{document}